\documentclass{article}

\usepackage[nonatbib,final]{neurips_2022}

\usepackage[utf8]{inputenc} 
\usepackage[T1]{fontenc}    
\usepackage{hyperref}       
\usepackage{url}            
\usepackage{booktabs}       
\usepackage{amsfonts}       
\usepackage{nicefrac}       
\usepackage{microtype}      
\usepackage{xcolor}         
\usepackage[pdftex]{graphicx}
\usepackage{lipsum}
\usepackage{amsmath}
\usepackage{subcaption}
\usepackage{caption}
\usepackage{capt-of}
\usepackage{wrapfig}

\usepackage[numbers]{natbib}

\title{Segmentation of Multiple Sclerosis Lesions across Hospitals: Learn Continually or Train from Scratch?}

\author{%
Enamundram Naga Karthik$^{1,2}$\thanks{Corresponding Author: \texttt{naga-karthik.enamundram@mila.quebec}} \quad 
Anne Kerbrat$^3$ \quad 
\textbf{Pierre Labauge}$^4$ \\
\textbf{Tobias Granberg}$^5$ \quad
\textbf{Jason Talbott}$^6$ \quad 
\textbf{Daniel S. Reich}$^7$ \quad
\textbf{Massimo Filippi}$^{8, 9}$ \\
\textbf{Rohit Bakshi}$^{10}$ \quad
\textbf{Virginie Callot}$^{11, 12}$ \quad
\textbf{Sarath Chandar} $^{2, 13}$ \quad 
\textbf{Julien Cohen-Adad}$^{1, 2, 14}$\\
$^1$NeuroPoly Lab, Polytechnique Montreal, Canada \quad 
$^2$MILA, Quebec AI Institute \\
$^3$Neurology Department, CHU Rennes, Rennes, France \\
$^4$MS Unit, Department of Neurology, CHU Montpellier, Montpellier, France \\
$^5$Department of Clinical Neuroscience, Karolinska Institutet, Stockholm, Sweden \\
$^6$Department of Radiology and Biomedical Imaging, Zuckerberg San Francisco \\
General Hospital, University of California, San Francisco, CA, USA \\
$^7$National Institute of Neurological Disorders and Stroke, National Institutes of Health, USA \\
$^8$Neuroimaging Research Unit, Institute of Experimental Neurology, Division of \\ Neuroscience, and Neurology Unit, IRCCS San
Raffaele Scientific Institute, Milan, Italy \\
$^{9}$Vita-Salute San Raffaele University, Milan, Italy\\
$^{10}$Brigham and Women’s Hospital, Harvard Medical School, Boston, USA \\
$^{11}$AP-HM, CHU Timone, Pole de Neurosciences Cliniques, Department of Neurology, \\ Marseille, France \quad $^{12}$Aix-Marseille Univ, CNRS, Marseille, France \\
$^{13}$Department of Computer and Software Engineering, Polytechnique Montreal, Canada \\
$^{14}$Functional Neuroimaging Unit, CRIUGM, Université de Montréal, Montréal, Canada 
}

\begin{document}

\maketitle

\begin{abstract}

Segmentation of Multiple Sclerosis (MS) lesions is a challenging problem. Several deep-learning-based methods have been proposed in recent years. However, most methods tend to be \textit{static}, that is, a single model trained on a large, specialized dataset, which does not generalize well. Instead, the model should learn across datasets arriving sequentially from different hospitals by building upon the characteristics of lesions in a \textit{continual} manner. In this regard, we explore experience replay, a well-known continual learning method, in the context of MS lesion segmentation across multi-contrast data from 8 different hospitals. Our experiments show that replay is able to achieve \textit{positive backward transfer} and reduce catastrophic forgetting compared to sequential fine-tuning. Furthermore, replay outperforms the multi-domain training, thereby emerging as a promising solution for the segmentation of MS lesions.
The code is available at \href{https://github.com/naga-karthik/continual-learning-ms}{this link}.

\end{abstract}

\section{Introduction}

Multiple Sclerosis (MS) is a chronic, neurodegenerative disease of the central nervous system.
Lesion segmentation from magnetic resonance images (MRI) serves as an important biomarker in measuring disease activity in MS patients. However, manual segmentation of MS lesions is a tedious process, hence motivating the need for automated tools for segmentation. 
Several deep-learning (DL) based methods
have been proposed in the past few years \citep{Zeng2020ReviewOD, larosa2020}. They tend to be trained in a \textit{static} manner - all the datasets are pooled, jointly preprocessed, shuffled (to ensure they are independent and identically distributed, IID) and then fed to the DL models. While this has its benefits, it does not represent a realistic scenario. First, it is difficult to pool datasets from multiple hospitals with increasing privacy concerns. Second, since MS is a chronic disease, one would imagine a scenario where a DL model, like humans, engages in continual learning (CL) \cite{Thrun1998} and builds upon the lesion characteristics from different centers when presented sequentially. However, this sequential knowledge acquisition 
presents a major problem in DL models known as \textit{catastrophic forgetting} \cite{McCloskey1989CatastrophicII}.      

\begin{figure}
\centering
\begin{minipage}[b]{0.3\textwidth}
\captionsetup{font=small}
  \caption{Overview of our methods. Four experiments were performed -  \textcolor{red}{A}: \textit{Single-domain training}: a model is trained individually on each center.
  \textcolor{red}{B}: \textit{Sequential fine-tuning}: after training the model on center \textit{n}, the pre-trained encoder weights are loaded for center \textit{n+1} (red dashed arrows). \textcolor{red}{C}: \textit{Experience replay}: in addition to fine-tuning (as in B) upto 20 samples per each center are stored in the memory buffer (in gray). \textcolor{red}{D}: \textit{Multi-domain training}: data from all centers are pooled and a joint model is trained. }
\label{fig: overview}
\vspace{-10pt}
\end{minipage}
\hspace{0.02\textwidth}
\begin{subfigure}[b]{0.65\textwidth}
  \includegraphics[width=1.0\linewidth]{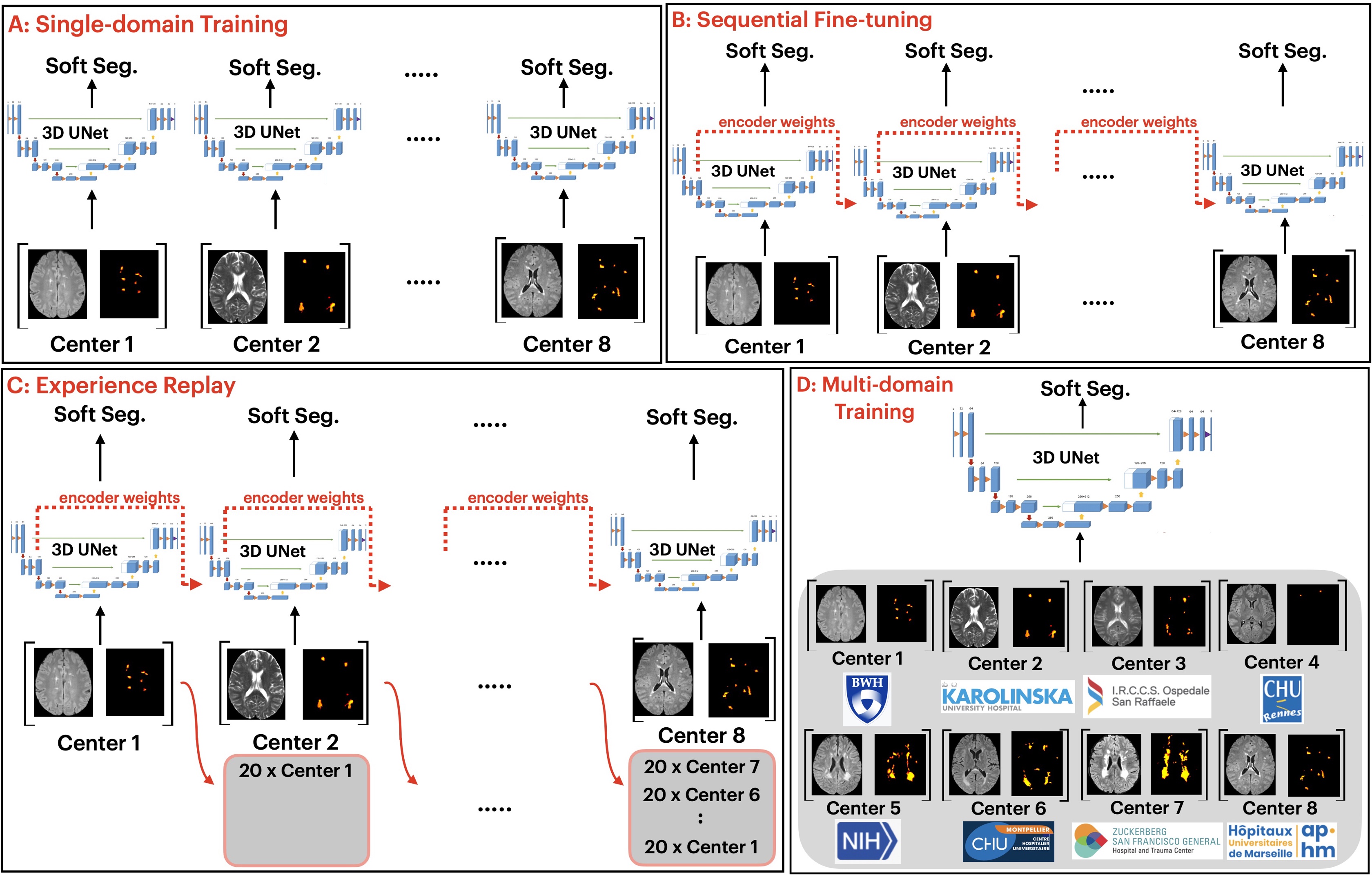}
\end{subfigure}
\end{figure}

Previous works in CL for medical imaging have used regularization-based \citep{Baweja2018TowardsCL, Garderen2019TowardsCL} and memory-based methods \cite{Hofmanninger2020DynamicMT} for tackling catastrophic forgetting. 
In this work, we formalize the MS lesion segmentation across multiple hospitals as a domain-incremental learning problem \cite{vandeVen2019ThreeSF}, where the task remains unique (i.e. segmentation of lesions) but the model is sequentially presented with new domains (i.e. data from different hospitals).
Four types of experiments are performed as shown in Figure \ref{fig: overview} - \textit{single-domain}, \textit{multi-domain} 
\textit{sequential fine-tuning}, and \textit{experience replay}. 
Our results show that replay helps in reducing catastrophic forgetting and achieves \textit{positive backward transfer}, that is, the segmentation performance on data seen earlier improves as the model continues to learn sequentially. 
Furthermore, we show that replay outperforms multi-domain training as more data arrive sequentially, thereby suggesting that the CL is a better long-term solution than re-training the model from scratch on a large, curated dataset.


\section{Experience Replay for MS Lesion Segmentation}
Replay (or, rehearsal) presents a straightforward way to prevent catastrophic forgetting and improve the performance on new domains. 
Let \textit{x} denote the patches of the 3D volumes, \textit{y} be the corresponding labels, $f_\theta$ denote the neural network with parameters $\theta$, and $\mathcal{D}$ denote the joint dataset. In a standard IID training regime, the loss $\mathcal{L}$ is given by Equation \ref{eq-loss_standard}.

\vspace{-8pt}
\noindent\begin{minipage}{.33\linewidth}
\begin{equation}
  \mathcal{L} = \mathbb{E}_{(x,y) \sim D} \left[ \ell(f_\theta (x), y) \right]
\label{eq-loss_standard}
\end{equation}
\end{minipage}%
\quad\begin{minipage}{.63\linewidth}
\begin{equation}
\mathcal{L'} = \mathbb{E}_{(x,y) \sim D_k} \left[ \ell (f_\theta (x), y) \right] + \mathbb{E}_{(x,y) \sim \mathcal{M}} \left[ \ell (f_\theta (x), y) \right]
\label{eq-loss_cl}
\end{equation}
\end{minipage}

In this work, we use the simplest form of experience replay wherein training data from previously encountered domains\footnote{We use \textit{domains} and \textit{centers} interchangeably. A domain is essentially a center (i.e. a hospital) that holds/provides the data. Hence, data from each new center is treated as a different domain.} are stored in a memory buffer and interleaved with the current domain's training data. Particularly, the dataset $\mathcal{D}$ is divided into 8 different domains ($\mathcal{D}_1, \mathcal{D}_2, \ldots \mathcal{D}_8$). The model is trained sequentially on one dataset $\mathcal{D}_k$ at a time. For each dataset $\mathcal{D}_k$ ($1 \leq k < 8$), we store upto 20 image-label pairs (depending on the dataset size) in a memory buffer $\mathcal{M}$ and merge them with the training data of the current domain. The updated loss term $\mathcal{L'}$ is given by Equation \ref{eq-loss_cl}.
Due to unconstrained access to the multi-center data, the model is tested on all the remaining centers once it has been trained on one center. We use the Dice Loss \cite{Milletari2016VNetFC} as the loss function $\ell$. 



\paragraph{SoftSeg}
To account for the partial volume effects at the lesion boundaries,
we use soft ground truth labels in our training procedure. In addition to mitigating the partial volume effects, soft segmentations \cite{Gros2021SoftSegAO} were shown to generalize better and to reduce model-overconfidence. In this regard, the notable changes include: (i) bypassing the binarization step after data preprocessing and augmentation, hence keeping the labels between $[0, 1]$, and (ii) using \textit{normalized ReLU} as the final activation function.

\section{Experiments and Results}


\paragraph{Data} We used the brain MRI datasets described in Kerbrat et al. \cite{kerbrat2020} containing 290 subjects from 8 different centers. We denote the centers with the following abbreviations along with their number of subjects: BWH: $n=80$, Karo: $n=51$, Milan: $n=47$, Rennes: $n=51$, NIH: $n=28$, Montp: $n=13$, UCSF: $n=12$, and AMU: $n=8$.
Six out of eight centers used 3D FLAIR scans, center \textit{Karo} used both 3D FLAIR and T2-weighted (T2w) scans, and center \textit{Milan} used only T2w scans. The data were pre-processed using the publicly-available \href{https://github.com/neuropoly/spinalcordtoolbox}{Spinal Cord Toolbox} and \href{https://github.com/Inria-Visages/Anima-Public}{Anima Toolbox}. The intra-subject registration between the $T_2$ and the FLAIR images was achieved using rigid transformations and subsequently registered to the ICBM template space. All the 3D MRI images were resampled to an isotropic $1 \textrm{mm}$ resolution. We refer the reader to \cite{kerbrat2020} for more details.

\paragraph{Experiments} 

A three-layer 3D UNet \cite{Ronneberger2015UNetCN} with residual connections was used. The data were split according to the 80/20 train/test ratio. For fine-tuning and replay experiments, the model observed each domain in sequence. The ordering of domains was randomly shuffled with 9 different seeds (more details in the supplementary material, Section \ref{subsec:hyperparams}).


\paragraph{Evaluation Metrics}
On a held-out test set, we used the Dice coefficient to evaluate the quality of the lesion segmentations and computed the backward transfer (BWT) metric \cite{LopezPaz2017GradientEM} to evaluate the CL capabilities of our model. 
Concretely, BWT quantifies the influence that training on center $n$ has on the performance on a previous center $ k < n$. Hence, a positive BWT occurs when the Dice score on a center $k < n$ \textit{increases} after training on center $n$ and vice-versa for a negative BWT. 

\begin{figure}[htbp!]
  \centering
\includegraphics[width=1.0\linewidth]{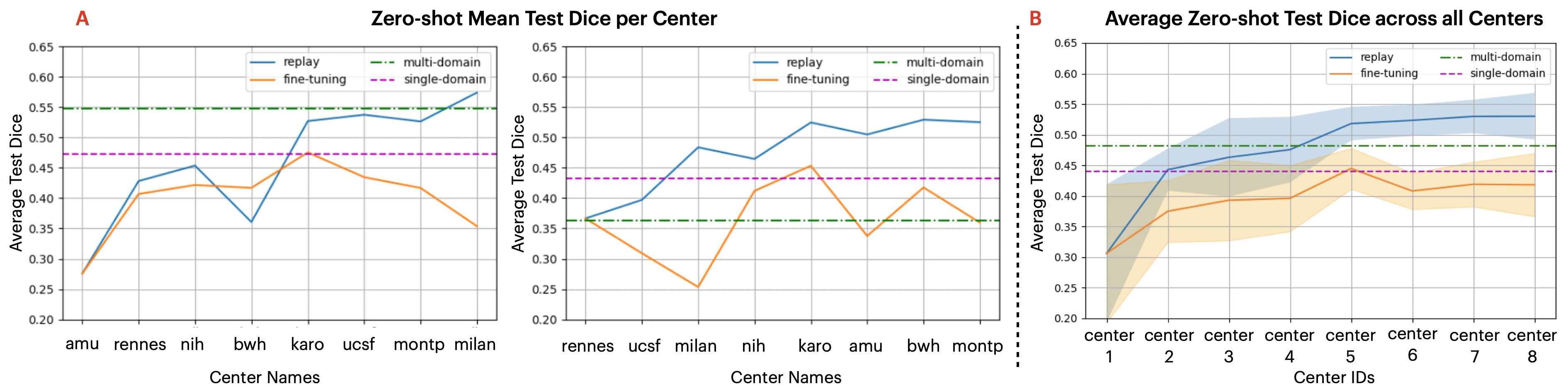}
\caption{Zero-shot (ZS) Test Dice scores with different random sequences of domains. \textcolor{red}{A}: ZS Test Dice scores with 2 random domain sequences. \textcolor{red}{B}: ZS Test Dice scores averaged across 9 randomly shuffled domain sequences.}
\label{fig:results}
\end{figure}
\begin{figure}[h]
\vspace{-15pt}
  \begin{minipage}[htbp!]{.65\linewidth}
    \centering
    \includegraphics[width=0.85\linewidth]{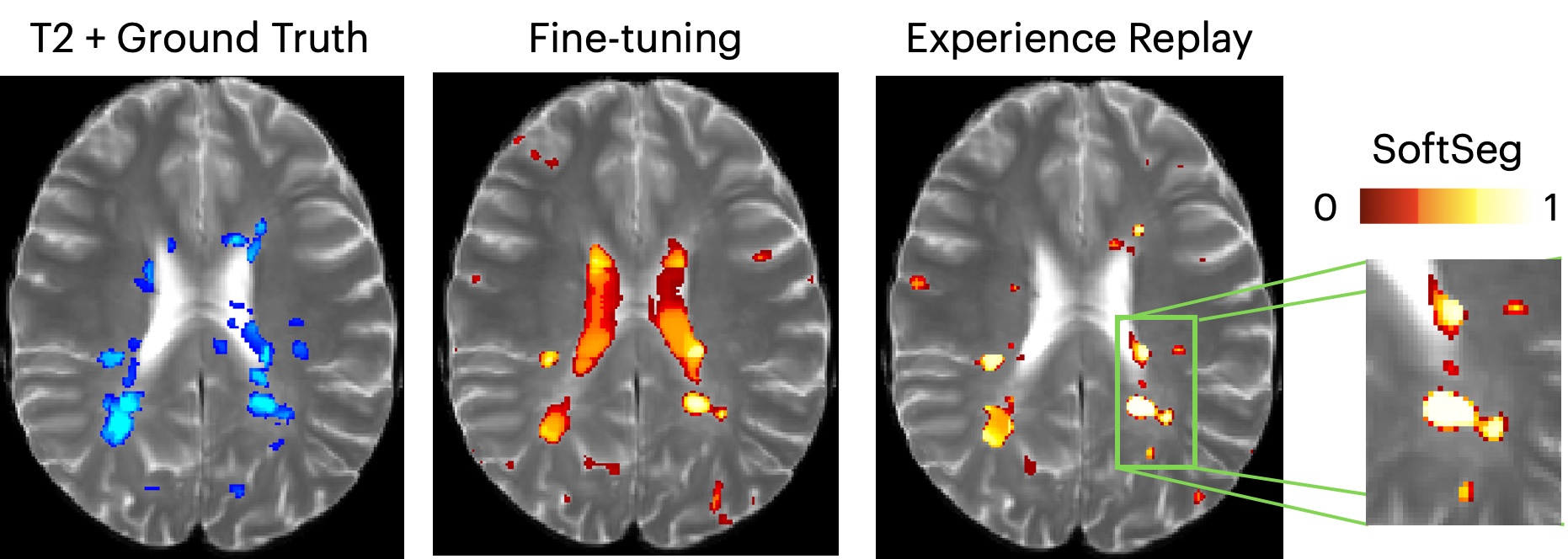}%
    \caption
      {%
        Qualitative results on a test sample from \textit{milan} center. Replay obtains better soft segmentations compared to fine-tuning.%
        \label{fig:soft-seg}%
      }%
  \end{minipage}\hspace{0.01\textwidth}
  \resizebox{0.32\linewidth}{!}{\begin{minipage}[htbp!]{.45\linewidth}
    \centering
    \captionof{table}
      {%
        BWT over descending order of domains (averaged across 9 seeds)%
        \label{tab:cl-results}%
      }
    \begin{tabular}{lccc}
    
    \toprule
    {\textbf{Center} $\downarrow$} &  \multicolumn{2}{c}{\textbf{Backward Transfer (BWT)}} \\
    
    \midrule
    {\textbf{Method} $\rightarrow$} & \textbf{Fine-tuning} & \textbf{Replay} \\ 
    \midrule
    BWH     & -0.07     &  \textbf{0.005} \\
    Karo    & -0.171    &   \textbf{0.037} \\
    Milan   & -0.284    &   \textbf{0.001}  \\
    Rennes  & -0.083    &   -0.012      \\
    NIH     & -0.119    &    -0.002     \\
    Montp   &  -0.061   &   \textbf{0.032} \\
    UCSF    &  -0.061   &   -0.008  \\
    AMU    & 0.0       &    0.0     \\
    \bottomrule
    \end{tabular}
  \end{minipage}}
\end{figure}

\paragraph{Results}
Figure \ref{fig:results}\textcolor{red}{A} shows the mean zero-shot test performance over 2 random sequences (more in section \ref{subsec:all-sequence-results}).
When learning to transfer knowledge across FLAIR $\to$ T2 contrasts, we observed large drops in Dice scores on center \textit{milan} as a consequence of catastrophic forgetting. On the other hand, replay performs better in this case and also exceeds the performance on multi-domain training. 
More importantly, Figure \ref{fig:results}\textcolor{red}{B} shows that over 9 random sequences of the domains, replay improves the segmentation performance over fine-tuning \textbf{and} multi-domain training as more data arrive. Thus, the proposed CL approach is capable of surpassing the implicit upper-bound defined by multi-domain (IID) training.  
In Figure \ref{fig:soft-seg}, we show the soft segmentations (ranging from $[0,1]$) obtained from fine-tuning and experience replay on a test sample from the \textit{milan} center. Fine-tuning incorrectly segments the periventricular region of the brain as lesions, whereas replay results in a better segmentation, while also providing a measure of uncertainty at the boundaries.

In Table \ref{tab:cl-results}, we report the BWT in terms of the test Dice scores on a fixed, descending order of the domains (defined as per the number of subjects). Large negative BWT was observed especially with centers \textit{karo} and \textit{milan}, implying catastrophic forgetting. 
On the other hand, not only does replay improve performance on these centers, it also achieves a \textit{positive BWT}, implying that training on the rest of the domains is indeed beneficial for these 2 domains.







\section{Conclusion}
This work presented a case for continual learning using experience replay for the segmentation of MS lesions.
Qualitative and quantitative results show that replay performs better than sequential fine-tuning in general, and especially when learning across contrasts FLAIR $\leftrightarrow$ T2. More importantly, it also outperforms multi-domain (IID) training as the data continue to arrive. 
Thus, storing a few samples per domain and rehearsing them regularly can improve performance over the long-term,
instead of re-training the model from scratch with each additional domain, which can be impractical.


\section{Potential Societal Impact}

\paragraph{Positive Societal Impact}


Our work presents interesting directions for future research. In particular, one could combine data from centers with very few subjects into a single center, thus making efficient use of the data available. Furthermore, new hospitals joining the existing group could be easily incorporated into the existing continual learning framework, instead of re-training the model from scratch on all the previous hospitals, thus saving a lot of computational overhead.

\paragraph{Negative Societal Impact}
Although relatively tiny, the environmental costs of training DL models do exist with our work, as with all the current DL models. A limitation is that the Dice scores reported are not yet suitable for clinical use. This might possibly be due to the sizes of the datasets from individual hospitals being relatively small.

\begin{ack}

We thank Pranshu Malviya for thoughtful suggestions and discussions. We also thank the physicians Pr. J. Pelletier, Pr. B. Audoin, and Dr. A. Maarouf from the Department of Neurology of La Timone Hospital, CHU Rennes for patient inclusions. ENK is supported by FRQNT and UNIQUE Excellence Doctoral scholarships.  


\end{ack}



\bibliographystyle{unsrtnat}
\bibliography{references}

\appendix

\section{Appendix}

In this section, we show the results from few additional experiments along with more details on the experiments and the hyperparameters used. 


\subsection{Experiment Details and Hyperparameter Settings}
\label{subsec:hyperparams}

All experiments were performed on an NVIDIA RTX A6000 GPU. For the computation times, an experiment with replay across all the 8 centers took about 11 hours, a fine-tuning experiment took about 5 hours, a multi-domain training took about 4.5 hours, and a single-domain experiment ranged between 10 mins to 1.5 hours (depending on the number of subjects per center). Higher training times for replay is explained by the inclusion of the memory buffer that accumulates samples as the training progresses. We ran our experiments on 9 different random seeds that were used for shuffling the order of the domains. All the results are averaged over these 9 seeds. We store upto 20 image-label pairs in the memory buffer per center. However, some centers with less data might not have 20 subjects. Hence, in those cases, we store the entire dataset for that center.
For the evaluation metric, we computed the Dice coefficient after binarizing the model's soft predictions with a threshold of $0.5$ for easier comparison with the other works.

As for the hyperparameters, we used a three-layer 3D UNet with 32 initial feature maps that are progressively doubled until the bottleneck is reached. The model was trained on isotropic 3D image patches of size $64 \times 64 \times 64$, with 4 weighted patches per image being sampled.  The AdamW optimizer was used with learning rate (LR) of $0.0001$ and a LR scheduler (stepLR) that decreased the LR by a factor of 0.5 every 50 epochs. We trained the model for 150 epochs with a batch-size of 4. 

\subsubsection{Sub-optimal hyperparameters}
We trained several models on different hyperparameters. We report some observations from the \textit{hyperparameters that did not work} to better inform the readers. Our main criteria for disregarding these hyperparameters include relatively smaller Dice scores and longer training times with minimal gains. 
In particular, training a 4-layer 3D UNet resulting in quick overfitting with higher training Dice scores and lower generalization. We believe that this is due to the small dataset sizes. Patch-sizes of $96 \times 96 \times 96$ achieved slightly higher test Dice scores but at the cost longer training times ($ \sim 1$ or $>1$ day for one replay experiment). On the other hand, patch-sizes of $32 \times 32 \times 32$ resulted in smaller Dice scores. 

\subsubsection{Calculating Backward Transfer (BWT)}
As mentioned earlier, after the model is trained on data from one domain, it is tested on all the $K=8$ domains. This helps in constructing a matrix $R \in \mathbb{R}^{K \times K}$, where $R_{\mathcal{D}_i, \mathcal{D}_j}$ corresponds to the test Dice score of the model after training on domain $\mathcal{D}_i$ and testing on domain $ \mathcal{D}_j$. Then, BWT can be calculated as:
\begin{equation*}
    \textrm{BWT} = \frac{1}{K-1} \sum_{i=1}^{K-1} R_{\mathcal{D}_K, \mathcal{D}_i} - R_{\mathcal{D}_i,\mathcal{D}_i}
\end{equation*}
where $\mathcal{D}_K$ refers to the data from the final domain/center in the sequence.

\subsection{Individual Results from Random Shuffling of Domains}
\label{subsec:all-sequence-results}

\begin{figure}[htbp!]
  \centering
\includegraphics[width=1.0\linewidth]{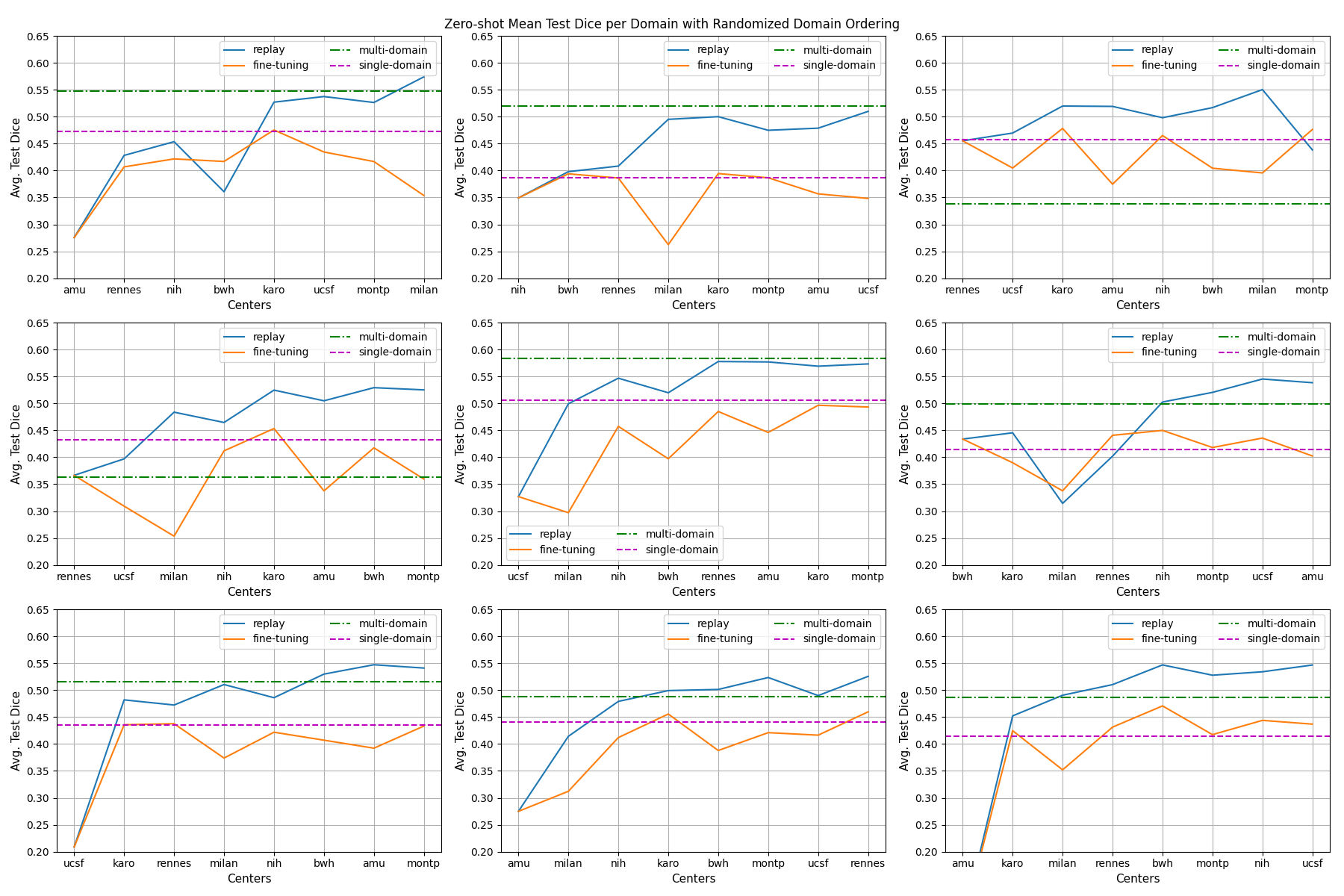}
  \caption{Comparison of the zero-shot test Dice scores of the four experiments on 9 random sequences of domains.}
\label{fig:zero-shot-fixed}
\end{figure}

Continuing from the main paper which compared the results of the 4 experiments on 2 different sequences of domains, Figure \ref{fig:zero-shot-fixed} shows the rest of the results on 7 additional random domain sequences. In every sequence, we observed that replay performed better than sequential fine-tuning and single-domain training. Interestingly, as more data arrive sequentially, replay matches or outperforms the multi-domain training, which is typically considered the upper-bound in terms of the performance. This suggests that replay could be a promising long-term solution of improving the segmentation performance across domains instead of creating a large dataset and training from scratch. 

\subsection{Results from the Fixing the Sequence of Domains}

\begin{figure}[htbp!]
  \centering
\includegraphics[width=1.0\linewidth]{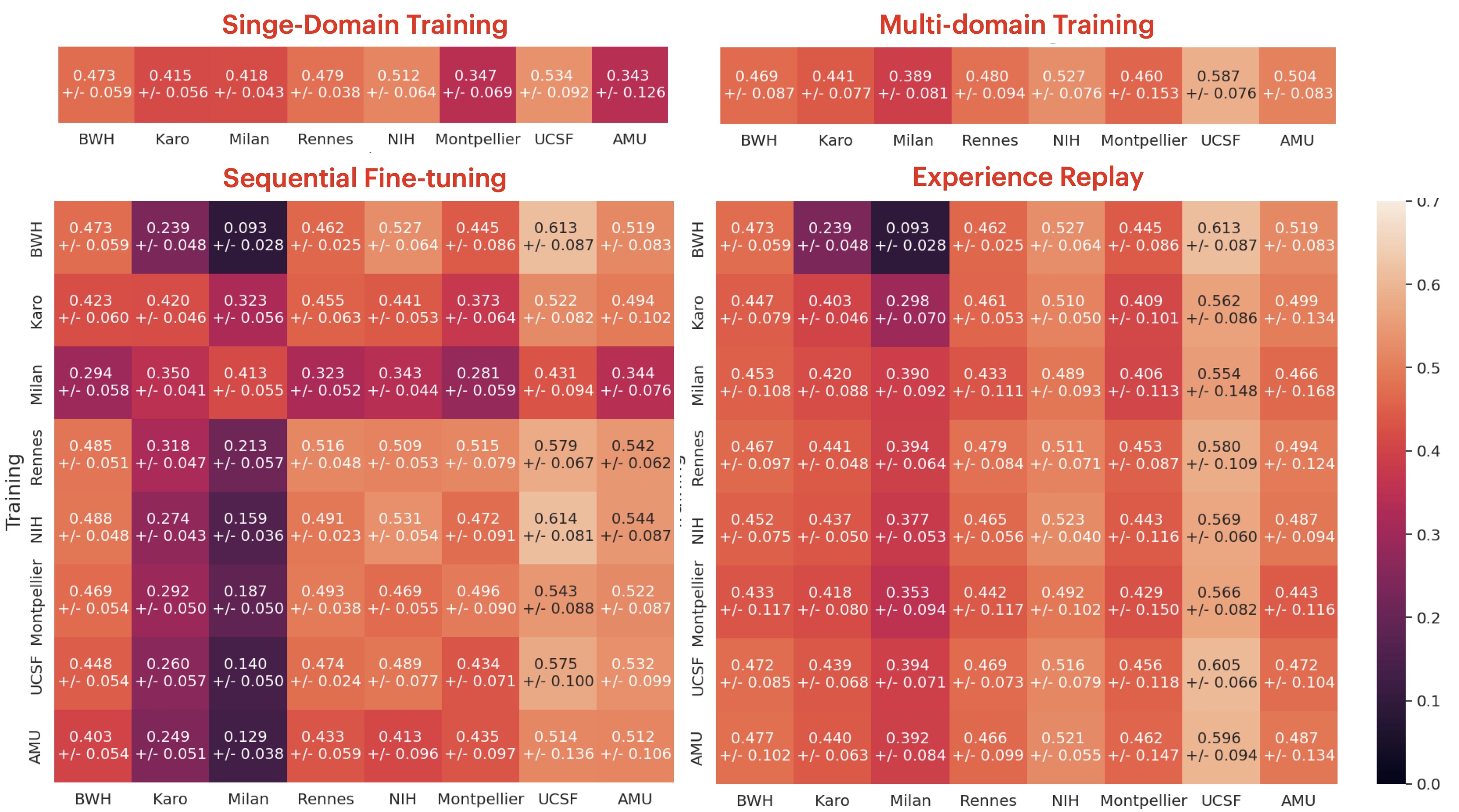}
  \caption{Comparison of Test Dice scores with fixed domain ordering across the 4 types of experiments. }
\label{fig:heatmap}
\end{figure}

In this experiment, we fixed the sequence of domains in a descending order according to the number of subjects per each center, resulting in the following sequence - [BWH, Karo, Milan, Rennes, NIH, Montp, UCSF, AMU] and trained the model on 9 different seeds. Figure \ref{fig:heatmap} shows the results (mean test Dice $\pm$ std.) from all the 4 types of experiments. The Y-axis represents the center the model was trained on and the X-axis represents the test performance on all the centers. For instance, in the $8 \times 8$ matrix for the fine-tuning and experience replay, element [2,4] shows the test Dice scores after training on center \textit{Milan} and testing on center \textit{NIH}. We observed that fine-tuning does not perform well when learning/transferring the knowledge across contrasts (seen in Row 3). Moreover, considering the 2nd and 3rd columns of the fine-tuning heatmap, it can be seen that even after reaching the final domain (\textit{AMU}), the performance on center \textit{Milan}, that contains T2w images, does not improve. This is different in the case of replay where training on data from other centers has helped in improving the performance on center \textit{Milan}, thereby implying a positive backward transfer.

\end{document}